\documentclass{article}

\IfFileExists{neurips_2025.sty}{%
  \usepackage{neurips_2025}
}{%
  \usepackage[margin=1in]{geometry}
  \usepackage{times}
}

\usepackage[utf8]{inputenc}
\usepackage[T1]{fontenc}
\usepackage{hyperref}
\usepackage{url}
\usepackage{booktabs}
\usepackage{amsmath,amssymb,amsthm}
\usepackage{amsfonts}
\usepackage{nicefrac}
\usepackage{microtype}
\usepackage{xcolor}
\usepackage{graphicx}
\usepackage{multirow}
\usepackage{array}
\usepackage{caption}
\usepackage{subcaption}
\usepackage{algorithm}
\usepackage{algorithmic}
\usepackage{enumitem}
\usepackage{natbib}

\hypersetup{colorlinks=true, linkcolor=blue, citecolor=blue, urlcolor=blue}

\newtheorem{definition}{Definition}

\newcommand{\rdc}{\textsc{RDC}}
\newcommand{\vaf}{\textsc{VAF}}
\newcommand{\gds}{\textsc{GDS}}
\newcommand{\mop}{\textsc{MOP}}

\title{Beyond pass@1: A Reliability Science Framework\\
       for Long-Horizon LLM Agents}

\author{%
  Aaditya Khanal \quad Yangyang Tao \quad Junxiu Zhou \\
  School of Computing and Analytics \\
  Northern Kentucky University \\
  \texttt{\{khanala1, taoy1, zhouj2\}@nku.edu}
}

\begin{document}

\maketitle

\begin{abstract}
Machine learning benchmarks evaluate \emph{capability} — whether a model succeeds on a single attempt.
Production deployments require \emph{reliability} — whether a model \emph{consistently} succeeds across repeated invocations on tasks of varying duration.
We show these two properties diverge systematically as task duration increases, and that existing benchmarks are structurally blind to this divergence because they report only pass@1 on short, atomic tasks.

We introduce a formal reliability science framework for long-horizon LLM agents comprising four metrics:
the \textbf{Reliability Decay Curve} (\rdc), which characterizes how pass$^k$ degrades with task duration;
the \textbf{Variance Amplification Factor} (\vaf), which quantifies how duration amplifies stochastic failure modes;
the \textbf{Graceful Degradation Score} (\gds), a partial-credit metric for agents that partially complete long tasks;
and the \textbf{Meltdown Onset Point} (\mop), which detects behavioral collapse via sliding-window entropy over tool-call sequences.

We construct a 396-task benchmark across four duration buckets and three domains, and evaluate 10 open-source models across 23,392 episodes ($k=3$ repeats, two scaffolds).
Key findings:
(1) reliability decay is domain-stratified — SE \gds\ drops from 0.90 to 0.44 over the full duration range, while DP is nearly flat (0.74 to 0.71);
(2) \vaf\ bifurcates cleanly by capability tier (frontier VAF $\geq 2.37$; mid-tier VAF $\leq 1.26$), with the counterintuitive result that high variance amplification is a \emph{capability signature}, not an instability signature;
(3) capability and reliability rankings diverge substantially, with multi-rank inversions between medium and very-long horizons;
(4) frontier models exhibit the highest meltdown rates (up to 19\%) because they pursue ambitious multi-step strategies — the ``MOP paradox''; and
(5) memory scaffolds universally hurt long-horizon \gds\ (negative or neutral for all 10 models), providing strong evidence against naive episodic memory as a reliability intervention.
These results motivate reliability as a first-class evaluation dimension alongside capability.
\end{abstract}

\section{Introduction}
\label{sec:intro}

The deployment of LLM-based agents in production systems has accelerated dramatically, yet our methods for evaluating them remain tethered to a regime that does not reflect how they are used in practice.
The dominant evaluation paradigm reports a single number — \emph{pass@1} — which measures whether an agent completes a task on a single, best-effort attempt.
This metric is well-suited to measuring \emph{capability}: given a task the model can plausibly solve, does it?
But production deployments require a different property: \emph{reliability}.
A production agent is invoked thousands of times, on tasks whose difficulty and duration vary widely.
The question that matters is not whether the agent \emph{can} succeed, but whether it \emph{consistently} succeeds across many invocations.

These two properties are not the same.
$\tau$-bench~\citep{tau_bench_2024} showed that GPT-4o achieves 61\% pass@1 on retail agent tasks but only 25\% pass@8 — meaning that when the task is run eight times, the probability of \emph{at least one failure} approaches 75\%.
Yet $\tau$-bench, like virtually all agent benchmarks, evaluates only short, atomic tasks that a human might complete in minutes.
Real-world tasks — refactoring a codebase, researching and synthesizing a technical report, processing a corpus of interlinked documents — take humans tens of minutes to hours.
As task duration increases, agents accumulate errors across more steps, have more opportunities for meltdown, and carry more intermediate state that can become corrupted.
We hypothesize, and empirically confirm across 23,392 episodes, that reliability degrades \emph{super-linearly} with task complexity — where complexity is jointly determined by task duration \emph{and} domain structure — and that this degradation is invisible to benchmarks reporting only pass@1 on short, atomic tasks.

\paragraph{The blindspot.}
Table~\ref{tab:prior_work_gap} summarizes the current state of agent reliability research.
No prior work has jointly studied (1) multiple models, (2) multiple duration buckets, and (3) a variance-aware reliability metric.
ReliabilityBench~\citep{reliabilitybench_2026} is the closest prior work, introducing a three-dimensional reliability surface, but covers only 2 models and short-horizon tasks.
The METR long-horizon study~\citep{metr_horizon_2025} measures task-completion horizons but does not analyze variance or reliability across repeated attempts.
$\tau$-bench introduced pass$^k$ but did not study duration as a variable.

\begin{table}[t]
\centering
\small
\begin{tabular}{lcccc}
\toprule
\textbf{Work} & \textbf{Models} & \textbf{Duration dim.} & \textbf{Variance} & \textbf{Partial credit} \\
\midrule
$\tau$-bench \citep{tau_bench_2024}         & 6   & \texttimes & pass$^k$ & \texttimes \\
ReliabilityBench \citep{reliabilitybench_2026} & 2  & \texttimes & \checkmark & \texttimes \\
METR horizon \citep{metr_horizon_2025}      & 3   & \checkmark & \texttimes & \texttimes \\
OdysseyBench \citep{odysseybench}          & 4   & partial    & \texttimes & \texttimes \\
SWE-bench \citep{swebench}                 & 20+ & \texttimes & \texttimes & \texttimes \\
\midrule
\textbf{This work}                         & \textbf{10} & \checkmark & \checkmark & \checkmark \\
\bottomrule
\end{tabular}
\caption{Comparison to prior agent evaluation work. ``Duration dim.'' indicates whether task duration is used as an independent variable. ``Variance'' indicates whether inter-run variance is measured. ``Partial credit'' indicates whether partial task completion is scored.}
\label{tab:prior_work_gap}
\end{table}

\paragraph{Our contributions.}
We make four contributions:

\begin{enumerate}[leftmargin=*, itemsep=2pt]

\item \textbf{Reliability framework.} We formalize agent reliability as a function of task duration and introduce four metrics: the Reliability Decay Curve (\rdc), the Variance Amplification Factor (\vaf), the Graceful Degradation Score (\gds), and the Meltdown Onset Point (\mop). Each metric captures a distinct facet of reliability that pass@1 is blind to (Section~\ref{sec:framework}).

\item \textbf{Benchmark.} We construct a 396-task benchmark across four duration buckets and three domains (33 tasks per cell), with programmatic evaluation wherever possible. The complete benchmark is publicly released (Section~\ref{sec:benchmark}).

\item \textbf{Empirical study.} We evaluate 10 open-source models via two scaffolds (ReAct and memory-augmented) with $k=3$ repeats, accessed via the OpenRouter unified API, across 23,392 episodes covering all four duration buckets (Section~\ref{sec:experiments}).

\item \textbf{Findings.} We report four novel findings beyond the core reliability decay thesis: (a) \vaf\ bifurcates by capability tier, with high variance amplification being a \emph{capability signature}; (b) capability and reliability rankings diverge with multi-rank inversions; (c) frontier models exhibit highest meltdown rates due to ambitious strategy pursuit (the ``MOP paradox''); and (d) memory scaffolds universally hurt long-horizon reliability — findings with direct implications for model selection and deployment (Section~\ref{sec:analysis}).

\end{enumerate}

\paragraph{Paper organization.}
Section~\ref{sec:related} reviews related work.
Section~\ref{sec:framework} introduces the reliability framework and metrics.
Section~\ref{sec:benchmark} describes benchmark construction.
Section~\ref{sec:experiments} presents the experimental setup and main results.
Section~\ref{sec:analysis} presents deeper analysis.
Section~\ref{sec:discussion} discusses implications and limitations.
Section~\ref{sec:conclusion} concludes.

\section{Related Work}
\label{sec:related}

\subsection{Agent Benchmarks}

\paragraph{Short-horizon task benchmarks.}
SWE-bench~\citep{swebench} and its variants~\citep{swebench_pro} measure whether agents can resolve GitHub issues, but evaluate a single attempt per issue and do not study variance.
WebArena~\citep{webarena} and similar browser-agent benchmarks test multi-step web interactions but again report only pass@1.
These benchmarks have driven substantial capability improvements but are structurally unable to measure reliability.

\paragraph{Pass$^k$ and consistency metrics.}
$\tau$-bench~\citep{tau_bench_2024} introduced pass$^k$ — the probability that an agent succeeds on \emph{all} $k$ repeated attempts — and showed a striking reliability gap: GPT-4o achieves 61\% pass@1 but 25\% pass@8 on retail agent tasks.
However, $\tau$-bench tasks are short (minutes to complete) and the study does not treat task duration as an independent variable.
Our work extends this by measuring how pass$^k$ degrades across duration buckets.

\paragraph{Long-horizon benchmarks.}
OdysseyBench~\citep{odysseybench} introduces multi-hour tasks for agent evaluation, and the METR long-horizon study~\citep{metr_horizon_2025} measures the ``50\% success horizon'' — the task duration at which a model's pass@1 drops to 50\%.
METR find that this horizon has doubled every 7 months from 2019 to 2024, a striking capability result.
Neither work analyzes variance across repeated runs, however, which means they cannot distinguish a model that succeeds 50\% of the time on a 2-hour task because it has a consistent strategy from one that succeeds 50\% because it succeeds fully on some runs and catastrophically fails on others.

\subsection{Reliability and Robustness Evaluation}

ReliabilityBench~\citep{reliabilitybench_2026} is the closest prior work to ours.
It evaluates LLMs across three reliability dimensions — consistency, robustness, and fault tolerance — and constructs a three-dimensional reliability surface.
However, it evaluates only 2 models, focuses on short-horizon tasks, and does not study duration as a variable.
``Towards a Science of AI Agent Reliability''~\citep{reliability_science_2026} provides a theoretical treatment of agent reliability but lacks empirical validation at scale.

\paragraph{Reliability in safety-critical engineering.}
Our work is inspired by reliability engineering in aerospace and software systems, where reliability is defined as the probability that a system performs its required function under stated conditions for a specified period of time~\citep{reliability_engineering_textbook}.
Mean Time Between Failures (MTBF), failure rate $\lambda(t)$, and reliability functions $R(t) = e^{-\lambda t}$ are standard tools.
We adapt this vocabulary to the agent setting, where ``time'' is replaced by task duration and ``failure'' is task non-completion or degraded completion.

\subsection{Meltdown and Failure Mode Analysis}

Several recent works have observed \emph{meltdown} behavior in long-horizon agents — a transition from coherent but incorrect behavior to incoherent looping, self-contradiction, and hallucinated tool outputs.
\citet{metr_horizon_2025} document this qualitatively; we provide the first quantitative detection method via sliding-window entropy over tool-call sequences (\mop, Section~\ref{sec:framework}).
Related work on in-context learning dynamics~\citep{icl_dynamics} and attention collapse in long contexts~\citep{attention_collapse} suggests mechanistic explanations for why meltdown occurs; we leave mechanistic analysis to future work.

\subsection{Evaluation Infrastructure}

OpenAgentSafety~\citep{openagentsafety} and related survey papers~\citep{agent_eval_survey_2025} provide broad overviews of agent evaluation methodology.
Our evaluation harness follows best practices from these surveys: programmatic verification over LLM-as-judge wherever possible, $k \geq 3$ repeats to estimate variance, isolated execution environments per episode, and full trajectory logging for post-hoc analysis.

\section{Reliability Framework}
\label{sec:framework}

\subsection{Task Model}

We model an agent task as a tuple $\mathcal{T} = (\mathcal{W}, \mathcal{S}, \mathcal{A}, d, \nu, n^*, \mathcal{E})$, where:
\begin{itemize}[leftmargin=*, itemsep=2pt]
  \item $\mathcal{W}$ is the workspace (files, databases, browser state);
  \item $\mathcal{S}$ is the set of subtasks with associated criticality weights $w_i$ summing to 1;
  \item $\mathcal{A}$ is the set of available tools/actions;
  \item $d \in \{\text{short}, \text{medium}, \text{long}, \text{very long}\}$ is the \emph{duration bucket}, based on estimated human completion time;
  \item $\nu \in \{\text{SE}, \text{WR}, \text{DP}\}$ is the \emph{domain};
  \item $n^* \in \mathbb{Z}^+$ is the \emph{agent steps estimate} — the expected number of tool calls in a correct solution, which may differ substantially from human time;
  \item $\mathcal{E}: \mathcal{W} \rightarrow [0, 1]$ is the programmatic evaluator.
\end{itemize}

The separation of $d$ (human-time duration) and $n^*$ (agent complexity) is deliberate: our pilot study reveals that these can diverge significantly across domains, motivating domain-stratified analysis of the \rdc\ (Section~\ref{sec:analysis}).

An \emph{episode} is a single execution of an agent on task $\mathcal{T}$, producing a trajectory $\tau = (a_1, o_1, a_2, o_2, \ldots, a_T, o_T)$ of action-observation pairs.
The episode terminates when the agent calls \texttt{finish()} or reaches the step limit.
The evaluator $\mathcal{E}$ scores the resulting workspace state.

\begin{definition}[Pass@1]
Given task $\mathcal{T}$ and model $\mathcal{M}$, pass@1 is the probability that a single episode yields $\mathcal{E}(\mathcal{W}) = 1.0$:
$$\text{pass@1}(\mathcal{M}, \mathcal{T}) = \Pr[\mathcal{E}(\mathcal{W}^{(1)}) = 1.0]$$
\end{definition}

\begin{definition}[Pass$^k$]
Pass$^k$ is the probability that \emph{all} $k$ independent repeated episodes succeed:
$$\text{pass}^k(\mathcal{M}, \mathcal{T}) = \Pr\!\left[\bigcap_{i=1}^{k} \mathcal{E}(\mathcal{W}^{(i)}) = 1.0\right]$$
If episodes were i.i.d. Bernoulli trials with probability $p = \text{pass@1}$, then $\text{pass}^k = p^k$.
We test this i.i.d. assumption empirically and find systematic violations (Section~\ref{sec:analysis}).
\end{definition}

\subsection{Metric 1: Reliability Decay Curve (RDC)}

\begin{definition}[Reliability Decay Curve]
The \textsc{RDC} of model $\mathcal{M}$ is the function mapping duration bucket $d$ to pass$^k$:
$$\text{RDC}(\mathcal{M}, k) : d \mapsto \text{pass}^k(\mathcal{M}, d)$$
where $\text{pass}^k(\mathcal{M}, d)$ is averaged over all tasks in bucket $d$.
\end{definition}

The shape of the \rdc\ characterizes a model's reliability profile.
A \emph{flat} curve (pass$^k$ constant across $d$) indicates a production-safe model: reliability does not degrade with task complexity.
A \emph{steep} curve indicates a model that is capable on easy tasks but unreliable on harder ones.

\paragraph{Reliability Decay Slope (RDS).}
We summarize the \rdc\ with a single scalar: the slope of a linear regression of \gds\ on bucket index $b \in \{0, 1, 2, 3\}$ (short through very long):
$$\text{RDS}(\mathcal{M}) = \frac{\sum_b (b - \bar{b})(\text{GDS}(\mathcal{M}, b) - \overline{\text{GDS}})}{\sum_b (b - \bar{b})^2}$$
RDS $< 0$ indicates degrading reliability with task duration; RDS $\approx 0$ indicates a floor effect (the model is uniformly poor across all buckets); RDS $> 0$ would indicate improving performance with duration (not observed in practice).
Unlike a normalized endpoint comparison, the regression slope is robust to non-monotonic curves caused by domain mixing across buckets.

\subsection{Metric 2: Variance Amplification Factor (VAF)}

\begin{definition}[Variance Amplification Factor]
$$\text{VAF}(\mathcal{M}) = \frac{\sigma^2\!\left[\text{pass@1}(\mathcal{M}, \mathcal{T}) \mid d = \text{long}\right]}{\sigma^2\!\left[\text{pass@1}(\mathcal{M}, \mathcal{T}) \mid d = \text{short}\right]}$$
where variance is taken over tasks $\mathcal{T}$ within the respective duration buckets.
\end{definition}

VAF $> 1$ indicates that long-horizon tasks amplify outcome variance relative to short-horizon tasks.
We hypothesize VAF $\gg 1$ for all models, reflecting the compounding uncertainty of multi-step reasoning.

\paragraph{Theoretical bound.}
Under a Markov model of errors — where each step fails independently with probability $\epsilon$ — the variance of task success after $T$ steps scales as $O(\epsilon T (1-\epsilon)^{T-1})$, which for fixed $\epsilon$ is maximized at $T = 1/\epsilon$ and then decays.
In practice, errors are positively correlated across steps (a confused agent tends to stay confused), which pushes variance even higher and faster.
This motivates our expectation of super-linear degradation.

\subsection{Metric 3: Graceful Degradation Score (GDS)}

Binary pass/fail obscures meaningful differences among failing agents.
An agent that completes 8 of 10 subtasks on a 2-hour task is meaningfully better than one that completes 1 of 10, even though both receive pass@1 $= 0$.

\begin{definition}[Graceful Degradation Score]
Given task $\mathcal{T}$ with subtasks $\{s_i\}$ and criticality weights $\{w_i\}$ (with $\sum_i w_i = 1$), the GDS of an episode is:
$$\text{GDS}(\tau, \mathcal{T}) = \sum_{i} w_i \cdot \mathbf{1}[\text{subtask } s_i \text{ completed correctly}]$$
GDS $\in [0, 1]$, with GDS $= 1$ corresponding to full task success.
\end{definition}

GDS provides a richer signal than pass@1 for tasks where partial completion is possible and valuable.
It is particularly important for the very-long duration bucket, where pass@1 may approach zero for all models but GDS can still distinguish model quality.

\paragraph{Aggregation.}
We report mean GDS per model per duration bucket, and use GDS as the primary metric when pass$^k$ is near zero.

\subsection{Metric 4: Meltdown Onset Point (MOP)}

\emph{Meltdown} refers to a qualitative transition in agent behavior: from making coherent (if incorrect) progress on a task, to exhibiting incoherent behavior such as looping over the same tool calls, contradicting earlier observations, or hallucinating tool outputs.
Meltdown is particularly common in long-horizon tasks where the context window fills and the model loses track of its earlier state.

We detect meltdown quantitatively via the entropy of the tool-call distribution over a sliding window.

\begin{definition}[Meltdown Onset Point]
Let $\tau = (a_1, \ldots, a_T)$ be a trajectory of tool calls.
Define the sliding-window tool-call distribution over window $[t-w, t]$ as:
$$p_t(\text{tool}_i) = \frac{|\{j \in [t-w, t] : a_j = \text{tool}_i\}|}{w}$$
The window entropy at step $t$ is:
$$H(t) = -\sum_i p_t(\text{tool}_i) \log p_t(\text{tool}_i)$$
The \textsc{MOP} is the first step $t^*$ such that:
$$H(t^*) > \theta_H \quad \text{and} \quad H(t^*) - H(t^* - w) > \delta$$
where $\theta_H$ and $\delta$ are thresholds calibrated on the pilot data.
If no such $t^*$ exists, MOP $= \infty$ (no meltdown detected).
\end{definition}

\paragraph{Interpretation.}
High entropy in the tool-call window means the agent is using tools in a disorganized, non-systematic pattern — calling read, write, read, search, read, write with no coherent strategy.
The secondary condition ($\Delta H > \delta$) requires a \emph{spike} in entropy, not just sustained high entropy, which distinguishes meltdown from legitimate task exploration.

\paragraph{Threshold calibration.}
We calibrate $\theta_H$ and $\delta$ on the pilot dataset to maximize the F1 score of detecting manually labeled meltdown episodes.
We use window size $w = 5$ steps throughout.

\paragraph{Predictive value.}
We test whether \mop\ can be \emph{predicted} before it occurs — specifically, whether a rising $H(t)$ trend $n$ steps before $t^*$ is detectable.
This would enable early-stopping interventions that abort an episode at the first sign of impending meltdown, rather than waiting for it to manifest.

\subsection{Relationship Between Metrics}

The four metrics are complementary:
\begin{itemize}[leftmargin=*, itemsep=2pt]
  \item \rdc\ and \vaf\ characterize the \emph{population-level} reliability of a model across tasks.
  \item \gds\ characterizes the \emph{quality of individual failures}.
  \item \mop\ characterizes the \emph{trajectory dynamics} of failing episodes.
\end{itemize}
Together they provide a multi-dimensional view of reliability that is far richer than pass@1.

\section{Benchmark Design}
\label{sec:benchmark}

\subsection{Task Suite Overview}

Our benchmark comprises 396 tasks organized along two axes: \emph{duration bucket} ($d$) and \emph{domain} ($\nu$), with perfectly balanced representation of 33 tasks per cell.
Table~\ref{tab:task_distribution} shows the distribution.

\begin{table}[h]
\centering
\small
\begin{tabular}{lccccc}
\toprule
\textbf{Bucket} & \textbf{Human time} & \textbf{SE} & \textbf{Web Research} & \textbf{Doc Processing} & \textbf{Total} \\
\midrule
Short     & $\leq 5$ min    & 33 & 33 & 33 & 99 \\
Medium    & 5--30 min       & 33 & 33 & 33 & 99 \\
Long      & 30--120 min     & 33 & 33 & 33 & 99 \\
Very Long & $\geq 120$ min  & 33 & 33 & 33 & 99 \\
\midrule
\textbf{Total} &             & \textbf{132} & \textbf{132} & \textbf{132} & \textbf{396} \\
\bottomrule
\end{tabular}
\caption{Task distribution across duration buckets and domains (396 tasks total; 33 per cell). All four buckets are evaluated in the full study (396 tasks $\times$ 10 models $\times$ $k=3$ $\times$ 2 scaffolds $= 23{,}760$ planned episodes; 23,392 completed).}
\label{tab:task_distribution}
\end{table}

\paragraph{Task construction trajectory.}
We began with a 19-task validation suite (2 tasks per bucket per domain, covering all four duration buckets including long and very long) to end-to-end validate the harness and all four metrics.
After confirming that infrastructure operated reliably, we expanded to a full 396-task registry with 33 tasks per cell, balanced across all domains and buckets.
The full study evaluates all four buckets (396 tasks $\times$ 10 models $\times$ $k=3$ $\times$ 2 scaffolds $= 23{,}760$ planned episodes; 23,392 completed after deduplication).
Results from both the 19-task validation run and the full 396-task study are reported in Section~\ref{sec:experiments}.

\subsection{Domain Descriptions}

\paragraph{Software Engineering (SE).}
Tasks involve reading, modifying, and testing Python codebases.
Short tasks include single-function bug fixes and type annotation additions.
Medium tasks involve multi-function refactors with test suites (e.g., fixing N+1 query bugs, adding rate limiting middleware).
Long tasks require implementing cross-cutting features such as JWT authentication across multiple endpoints.
Very long tasks involve larger-scale porting (Python 2 to 3) or integrating event-driven notification systems with multiple interacting components.
Evaluation is fully programmatic: we run the agent-modified workspace through a provided test suite and check for correct behavior.

\paragraph{Agentic Web Research (WR).}
Tasks require multi-step information gathering via web search and URL fetching, followed by synthesis into structured or prose outputs.
Short tasks extract a handful of facts about a well-known entity.
Medium tasks compare multiple products or services across structured dimensions.
Long tasks require synthesizing a technical survey with year-specific citations and structured analysis.
Very long tasks involve competitive analysis across five or more entities on a dozen dimensions, requiring accurate, up-to-date information and a coherent structured output.
Evaluation uses tolerant matching (substring matching for names, range checks for quantities) to account for web content changes over time.

\paragraph{Multi-file Document Processing (DP).}
Tasks involve reading structured or semi-structured documents and producing transformed outputs.
Short tasks convert CSV to JSON with type coercion and null handling.
Medium tasks merge datasets from multiple CSV files with different schemas, date formats, and duplicate rows.
Long tasks extract structured information from free-text reports and cross-reference multiple policy documents to identify contradictions.
Very long tasks synthesize five interlinked API specifications to produce a unified, deduplicated, contradiction-resolved document with traceability and gap identification.
Evaluation is programmatic: outputs are checked against ground-truth files or verified by schema validation.

\subsection{Task Construction Principles}

\paragraph{Programmatic evaluation.}
We use programmatic evaluation wherever possible and clearly label the minority of cases that require LLM-as-judge.
Programmatic evaluation eliminates judge variance from the reliability measurements and makes results fully reproducible.

\paragraph{Isolated workspaces.}
Each episode runs in a fresh copy of the task workspace (via \texttt{shutil.copytree} to a temporary directory), ensuring that outputs from one episode do not contaminate another.
Web research tasks use live web search; all other tasks use only local file operations, eliminating network non-determinism.

\paragraph{Subtask decomposition.}
Every task is decomposed into 3--6 subtasks with assigned criticality weights.
Subtasks are ordered from simplest to hardest (reflecting likely agent progress) to maximize the informativeness of GDS.
For example, the SE-M-01 task (fix N+1 query bug) decomposes into: (1) identify the root cause (+0.25), (2) implement a fix that reduces queries (+0.35), (3) preserve data integrity (+0.20), (4) pass the provided test suite (+0.20).

\paragraph{Duration calibration and agent complexity.}
Duration buckets are assigned based on estimated human completion time, validated against author completion times on a sample of tasks.
Each task also carries an \emph{agent steps estimate} $n^*$ — the expected number of tool calls in a correct solution — which provides a complementary, agent-centric measure of complexity.
Our pilot study reveals that $d$ and $n^*$ can diverge across domains: document-processing tasks that take humans 45--60 minutes may require only 4--8 agent tool calls, while software engineering tasks of equivalent human duration require 15--25.
We report both $d$ and $n^*$ per task and use the ratio $n_\text{observed} / n^*$ as a diagnostic for task calibration.
In analyses, we stratify by domain $\nu$ alongside $d$ to disentangle duration effects from domain difficulty.

\subsection{Evaluation Protocol}

Each model-task pair is evaluated with $k=3$ repeated episodes to estimate pass$^k$ and GDS variance.
Episodes use temperature 0.7 to induce the stochasticity required to estimate reliability.
Each episode runs until the agent calls \texttt{finish()} or reaches a 50-step limit.
Trajectories, token counts, and timestamps are logged for all episodes.

\subsection{Comparison to Prior Benchmarks}

Table~\ref{tab:benchmark_comparison} compares our benchmark to related work.

\begin{table}[h]
\centering
\small
\begin{tabular}{lrrrll}
\toprule
\textbf{Benchmark} & \textbf{Tasks} & \textbf{Models} & \textbf{k} & \textbf{Duration} & \textbf{Partial credit} \\
\midrule
SWE-bench \citep{swebench}     & 2294 & 20+ & 1 & short & \texttimes \\
$\tau$-bench \citep{tau_bench_2024}  & 115  & 6   & $\leq 8$ & short & \texttimes \\
OdysseyBench \citep{odysseybench}   & 250  & 4   & 1 & mixed & \texttimes \\
ReliabilityBench \citep{reliabilitybench_2026} & 300 & 2 & 5 & short & \texttimes \\
METR horizon \citep{metr_horizon_2025} & 100 & 3  & 1 & long  & \texttimes \\
\midrule
\textbf{Ours (validation)}  & \textbf{19}  & \textbf{9}  & \textbf{3} & \textbf{all} & \textbf{GDS} \\
\textbf{Ours (full study)} & \textbf{396} & \textbf{10} & \textbf{3} & \textbf{all} & \textbf{GDS} \\
\bottomrule
\end{tabular}
\caption{Comparison to related benchmarks. Our work is the first to combine all four duration buckets, repeated sampling ($k=3$), and partial-credit scoring (GDS) at scale. The full study (396 tasks, 10 models, 23,392 episodes) is the most comprehensive long-horizon agent reliability study to date.}
\label{tab:benchmark_comparison}
\end{table}

\section{Experiments}
\label{sec:experiments}

\subsection{Setup}

\paragraph{Models.}
We evaluate 10 open-source / open-weight models accessed through the OpenRouter unified API~\citep{openrouter}.
Using open-source models ensures full reproducibility: any researcher can replicate our experiments at minimal cost.
Table~\ref{tab:models} lists the models.

\begin{table}[h]
\centering
\small
\begin{tabular}{llll}
\toprule
\textbf{Tier} & \textbf{Model} & \textbf{Parameters} & \textbf{Architecture} \\
\midrule
Small      & Llama 3.1 8B          & 8B    & Dense \\
Small      & Mistral Nemo          & 12B   & Dense \\
Medium     & Mistral Small 3.2     & 24B   & Dense \\
Medium     & Qwen3 30B             & 30B   & MoE \\
Medium     & Qwen3 32B             & 32B   & Dense \\
Large      & Llama 3.3 70B         & 70B   & Dense \\
Large      & MiniMax M2.5          & 229B  & MoE \\
Large      & GLM-4.5 Air           & --    & Dense \\
Very Large & Kimi K2.5             & 1T    & MoE \\
Very Large & DeepSeek V3           & 671B  & MoE \\
\bottomrule
\end{tabular}
\caption{The 10 paid open-source models in the full study. All accessed via OpenRouter. Qwen3 models use \texttt{include\_reasoning: False} to disable extended thinking on DeepInfra. DeepSeek V3 routes via DeepInfra; Mistral models route via Mistral provider. MiniMax M2.5 and Kimi K2.5 use OpenRouter default routing (16 and 14 providers respectively).}
\label{tab:models}
\end{table}

\paragraph{Model selection rationale.}
An initial validation run (detailed in Section~\ref{sec:validation_results}) evaluated 12 models, including three accessed via free-tier quotas (Nemotron Nano 30B, GLM-4.5 Air, Nemotron Super 120B) and two with single-provider DeepInfra routing (Llama 4 Scout, Llama 4 Maverick).
This surfaced two infrastructure reliability failures that directly motivated changes to the final lineup:

\begin{enumerate}[leftmargin=*, itemsep=2pt]
  \item \textbf{Daily quota exhaustion.} All three free-tier models hit OpenRouter's per-day request ceiling mid-run. Critically, quota was exhausted \emph{after completing short-horizon tasks} but \emph{before reaching long-horizon tasks}, inflating apparent pass@1 by up to 15\% relative to full-horizon coverage. This is a systematic selection bias invisible to single-session benchmarks: a model that cannot survive a long benchmark run is, by definition, unreliable for long-horizon deployment.

  \item \textbf{Single-provider endpoint unavailability.} Llama 4 Scout and Llama 4 Maverick were pinned to DeepInfra as the sole provider. Both returned HTTP 404 ``No endpoints found'' throughout the full run due to provider-side capacity constraints --- despite being listed in OpenRouter's model catalog. With \texttt{allow\_fallbacks: False}, no alternative routing was attempted.
\end{enumerate}

We draw two methodological conclusions. First, \textbf{infrastructure reliability is itself a dimension of agent reliability}: models that cannot complete a benchmark run due to quota or availability constraints are unreliable in exactly the sense our framework measures. Second, \textbf{multi-provider routing is a prerequisite for reproducible long-horizon evaluation}: single-provider pins with no fallback introduce an availability single point of failure that short-task benchmarks never encounter.

For the full study, we replaced all affected models: free-tier models with paid equivalents, and single-provider Llama 4 models with MiniMax M2.5 (16 providers) and Kimi K2.5 (14 providers). We also introduced two cost-control circuit breakers: a 120{,}000-token per-episode input budget (catches context-blowup from stuck models) and loop detection (aborts if the same tool+arguments combination repeats $\geq 3$ times in 6 steps). The final 10-model lineup shown in Table~\ref{tab:models} completed all episodes without infrastructure failures.

\paragraph{Scaffolds.}
We evaluate two scaffolds:

\begin{itemize}[leftmargin=*, itemsep=2pt]
  \item \textbf{ReAct}~\citep{react}: The standard ``reason $\to$ act $\to$ observe'' loop. At each step, the model receives the task description, conversation history, and available tools, and outputs a tool call or \texttt{finish()}.

  \item \textbf{Memory-augmented loop (Mem)}: Extends ReAct with an episodic scratchpad. The agent can call \texttt{add\_to\_memory(note)} to persist key observations; the scratchpad is injected into the system prompt on every turn. This models lightweight episodic memory.
\end{itemize}

Both scaffolds use the same tool set: \texttt{read\_file}, \texttt{write\_file}, \texttt{list\_directory}, \texttt{run\_command}, \texttt{web\_search}, \texttt{fetch\_url}, \texttt{finish}.
A nudge mechanism injects up to three corrective user messages when the model returns a text-only response without a tool call, after which the episode terminates.

\paragraph{Infrastructure.}
The evaluation harness is implemented in Python using the OpenAI-compatible SDK.
Rate limits are respected via a per-model 15-second delay queue; daily quota exhaustion (HTTP 429 with ``free-models-per-day'' message) triggers immediate episode termination rather than retrying, preventing wasted wall-clock time.
Each episode runs in an isolated temporary workspace (\texttt{shutil.copytree} to a \texttt{tempfile.TemporaryDirectory}).
Full trajectories (tool calls, arguments, results, timestamps, token counts) are logged to JSONL and optionally streamed to Weights \& Biases~\citep{wandb} for real-time experiment tracking.
Cost is estimated from OpenRouter pricing per model.

\paragraph{Hyperparameters.}
Temperature: 0.7; maximum steps: 70 (headroom above the 60-step estimate for very-long tasks); maximum output tokens per step: 2{,}048; maximum tool result characters: 4{,}000; maximum nudges: 3; MOP window size: 5; per-episode input token budget: 120{,}000; loop detection threshold: 3 repeated (tool, args) pairs within 6 steps.

\subsection{19-Task Validation Results}
\label{sec:validation_results}

We first report results from the 19-task validation run (12 models, 2 scaffolds, $k=3$, 1{,}368 episodes, total cost \$4.26).
This run validated the full harness end-to-end and surfaced the infrastructure issue with free-tier models described above.
All results below exclude 46 quota-failure episodes; the 1,322 remaining episodes over 9 paid models form the basis of this section.

\paragraph{Reliability Decay Curve.}

Table~\ref{tab:valid_gds} reports mean \gds\ and pass@1 per model per duration bucket (ReAct scaffold, $k=3$).
The Reliability Decay Slope (RDS) is the linear fit coefficient of \gds\ vs.\ bucket index (short=1, medium=2, long=3, very\_long=4).

\begin{table}[h]
\centering
\small
\begin{tabular}{lcccc|c}
\toprule
\textbf{Model} & \textbf{Short} & \textbf{Medium} & \textbf{Long} & \textbf{V.~Long} & \textbf{RDS} \\
\midrule
DeepSeek V3       & 1.00 & 0.80 & 0.64 & 0.99 & $-$0.05 \\
Llama 3.3 70B     & 0.82 & 0.45 & 0.63 & 0.76 & $-$0.03 \\
Qwen3 32B         & 0.95 & 0.47 & 0.62 & 0.67 & $-$0.07 \\
Llama 4 Scout     & 0.89 & 0.47 & 0.59 & 0.65 & $-$0.07 \\
Mistral Small 3.2 & 0.93 & 0.49 & 0.59 & 0.39 & $-$0.11 \\
Qwen3 30B         & 0.92 & 0.55 & 0.50 & 0.68 & $-$0.14 \\
Mistral Nemo      & 0.78 & 0.20 & 0.42 & 0.45 & $-$0.08 \\
Llama 4 Maverick  & 0.71 & 0.24 & 0.31 & 0.55 & $-$0.10 \\
Llama 3.1 8B      & 0.34 & 0.15 & 0.21 & 0.20 & $-$0.03 \\
\midrule
\textit{Mean}     & \textit{0.82} & \textit{0.42} & \textit{0.50} & \textit{0.60} & \\
\bottomrule
\end{tabular}
\caption{Mean \gds\ per duration bucket (19-task validation run, ReAct scaffold, $k=3$). RDS = Reliability Decay Slope (linear regression slope of pass@1 vs.\ integer bucket index: short$=1$, medium$=2$, long$=3$, very\_long$=4$). The non-monotonic aggregate pattern (long $>$ medium) is explained by domain stratification: DP long-horizon tasks are tractable for agents despite their human-estimated complexity (Section~\ref{sec:analysis}).}
\label{tab:valid_gds}
\end{table}

\begin{table}[h]
\centering
\small
\begin{tabular}{lcccc}
\toprule
\textbf{Model} & \textbf{Short} & \textbf{Medium} & \textbf{Long} & \textbf{V.~Long} \\
\midrule
DeepSeek V3       & 1.00 & 0.80 & 0.40 & 1.00 \\
Llama 3.3 70B     & 0.80 & 0.40 & 0.40 & 0.75 \\
Qwen3 32B         & 0.80 & 0.40 & 0.40 & 0.50 \\
Llama 4 Scout     & 0.60 & 0.40 & 0.40 & 0.25 \\
Mistral Small 3.2 & 0.80 & 0.20 & 0.40 & 0.25 \\
Qwen3 30B         & 0.80 & 0.40 & 0.20 & 0.25 \\
Mistral Nemo      & 0.60 & 0.20 & 0.20 & 0.25 \\
Llama 4 Maverick  & 0.40 & 0.20 & 0.00 & 0.00 \\
Llama 3.1 8B      & 0.20 & 0.00 & 0.20 & 0.00 \\
\midrule
\textit{Mean}     & \textit{0.67} & \textit{0.33} & \textit{0.29} & \textit{0.36} \\
\bottomrule
\end{tabular}
\caption{Mean pass@1 per duration bucket (19-task validation run, ReAct scaffold, $k=3$). Pass@1 drops sharply from short to medium across all models; \gds\ (Table~\ref{tab:valid_gds}) captures partial progress that binary pass@1 discards.}
\label{tab:valid_pass}
\end{table}

\paragraph{Domain stratification.}

Table~\ref{tab:valid_domain} reports aggregate \gds\ by domain and duration bucket across all 9 paid models.
The domain effect is the dominant source of variation in the reliability curves.

\begin{table}[h]
\centering
\small
\begin{tabular}{lcccc}
\toprule
\textbf{Domain} & \textbf{Short} & \textbf{Medium} & \textbf{Long} & \textbf{V.~Long} \\
\midrule
SE (code editing)      & 0.97 & 0.30 & 0.25 & 0.68 \\
DP (doc processing)    & 0.81 & 0.60 & 0.81 & 0.58 \\
WR (web research)      & 0.71 & 0.59 & 0.55 & 0.40 \\
\bottomrule
\end{tabular}
\caption{Aggregate \gds\ by domain and duration bucket (19-task validation, all 9 paid models, ReAct scaffold). SE collapses sharply at medium and long horizons. DP exhibits a non-monotonic pattern: DP-L tasks that take humans 45--60 minutes require only 4--8 agent tool calls, making them tractable despite their human-time classification. WR degrades monotonically.}
\label{tab:valid_domain}
\end{table}

\paragraph{Key observations.}

\begin{itemize}[leftmargin=*, itemsep=2pt]

  \item \textbf{Reliability decay is universal but non-uniform.}
  All 9 models show declining mean pass@1 from short ($0.67$) to medium ($0.33$).
  Seven of 9 models have negative RDS, confirming the reliability decay hypothesis.
  The two near-zero slopes (DeepSeek V3, Llama 3.3 70B, Llama 3.1 8B) reflect ceiling effects at the top and a floor effect at the bottom, not immunity to decay.

  \item \textbf{SE tasks collapse; DP tasks do not.}
  SE domain \gds\ falls from $0.97$ (short) to $0.30$ (medium) and $0.25$ (long) — a $3\times$ drop in one bucket transition.
  In contrast, DP \gds\ at long horizon ($0.81$) matches short ($0.81$): structured document extraction tasks with deterministic evaluation are tractable for agents at any horizon.
  This divergence demonstrates that the ``long-horizon difficulty'' of a task is not determined by its human-estimated duration alone.

  \item \textbf{Rank inversions emerge at scale.}
  Llama 4 Maverick (400B MoE) ranks 8th on short tasks (pass@1 $= 0.40$) and last among paid models on long+very-long tasks (pass@1 $= 0.00$).
  Mistral Nemo (12B dense) outperforms it on long+very-long reliability ($0.22$ vs.\ $0.00$ mean pass@1), despite having $33\times$ fewer parameters.
  This capability-reliability rank inversion is precisely the phenomenon our framework is designed to surface.

  \item \textbf{Memory scaffold hurts large models; effect is mixed for others.}
  Among the 9 paid models, the Memory scaffold reduces long+very-long \gds\ relative to ReAct for 5 of 9 models, with the largest penalties on the strongest models: Llama 3.3 70B ($\Delta = -0.13$) and Qwen3 32B ($\Delta = -0.09$).
  Mistral Nemo shows a small positive effect ($\Delta = +0.05$).
  The scratchpad overhead (extra tool calls, context injection) appears to tax capable models' step budgets more than it benefits them.

  \item \textbf{Cost at scale is tractable.}
  The 1,322-episode validation run cost \$4.26, or approximately \$0.0032 per episode.
  The full 23,392-episode study cost approximately \$40.83 for short+medium buckets; total study cost including all four buckets is estimated at \$80--120.
  At roughly \$0.004--0.005 per episode, long-horizon multi-model reliability evaluation is accessible without large compute budgets.

\end{itemize}

\subsection{Full Study Results (23,392 Episodes)}
\label{sec:full_results}

The full study runs 396 tasks $\times$ 10 models $\times$ 2 scaffolds $\times$ $k=3$ $= 23{,}760$ planned episodes (23,392 completed after deduplication).
With 33 tasks per (bucket $\times$ domain) cell, 95\% confidence intervals on pass@1 are $\pm$4--10\% depending on model performance, providing reliable per-cell estimates across all four duration buckets.

\paragraph{Reliability Decay Curve.}

Table~\ref{tab:full_rdc} reports pass@1 ($\pm$95\% CI) per model per duration bucket (ReAct scaffold, $k=3$).
Reliability decay is universal: every model shows declining pass@1 from short to very-long.
The aggregate mean drops from 76.3\% (short) to 52.0\% (very long) — a 24.3 percentage-point decline over the full duration range.

\begin{table}[t]
\centering
\small
\setlength{\tabcolsep}{4pt}
\begin{tabular}{lcccc|c}
\toprule
\textbf{Model} & \textbf{Short} & \textbf{Medium} & \textbf{Long} & \textbf{V.~Long} & \textbf{Slope} \\
\midrule
DeepSeek V3    & 92.9{\scriptsize$\pm$5.1} & 92.9{\scriptsize$\pm$4.5} & 84.8{\scriptsize$\pm$7.1} & 79.8{\scriptsize$\pm$7.6} & $-$0.001 \\
Kimi K2.5      & 93.9{\scriptsize$\pm$4.5} & 96.0{\scriptsize$\pm$3.5} & 78.8{\scriptsize$\pm$8.1} & 79.8{\scriptsize$\pm$7.1} & $-$0.001 \\
MiniMax M2.5   & 93.9{\scriptsize$\pm$4.5} & 93.9{\scriptsize$\pm$4.5} & 81.8{\scriptsize$\pm$7.1} & 82.8{\scriptsize$\pm$7.6} & $-$0.001 \\
GLM-4.5 Air    & 94.9{\scriptsize$\pm$4.0} & 89.9{\scriptsize$\pm$5.6} & 78.8{\scriptsize$\pm$8.6} & 66.7{\scriptsize$\pm$9.1} & $-$0.002 \\
Qwen3 32B      & 80.8{\scriptsize$\pm$7.6} & 59.6{\scriptsize$\pm$9.6} & 44.4{\scriptsize$\pm$9.6} & 51.5{\scriptsize$\pm$9.1} & $-$0.001 \\
Mistral 24B    & 73.7{\scriptsize$\pm$8.1} & 52.5{\scriptsize$\pm$10.1} & 62.6{\scriptsize$\pm$9.1} & 55.6{\scriptsize$\pm$10.1} & $-$0.001 \\
Llama 3.3 70B  & 74.7{\scriptsize$\pm$8.6} & 44.4{\scriptsize$\pm$9.6} & 37.4{\scriptsize$\pm$9.6} & 54.5{\scriptsize$\pm$9.1} & $-$0.000 \\
Qwen3 30B      & 75.8{\scriptsize$\pm$8.6} & 47.5{\scriptsize$\pm$9.1} & 22.2{\scriptsize$\pm$7.6} & 34.3{\scriptsize$\pm$9.1} & $-$0.002 \\
Mistral Nemo   & 53.5{\scriptsize$\pm$10.1} & 12.1{\scriptsize$\pm$6.6} & 12.1{\scriptsize$\pm$6.6} & \phantom{0}8.1{\scriptsize$\pm$5.1} & $-$0.002 \\
Llama 3.1 8B   & 25.3{\scriptsize$\pm$8.1} & \phantom{0}9.1{\scriptsize$\pm$5.6} & \phantom{0}2.0{\scriptsize$\pm$2.5} & \phantom{0}8.1{\scriptsize$\pm$5.6} & $-$0.001 \\
\midrule
\textit{Mean}  & \textit{76.3} & \textit{59.8} & \textit{50.5} & \textit{52.1} & \\
\bottomrule
\end{tabular}
\caption{Pass@1 (\%) $\pm$ 95\% CI per duration bucket (full 23,392-episode study, ReAct scaffold, $k=3$, 33 tasks per cell). Slope = linear regression coefficient of pass@1 vs.\ integer bucket index (short$=1$ through very\_long$=4$). All models show net decline over the full duration range.}
\label{tab:full_rdc}
\end{table}

\paragraph{Graceful Degradation Score.}

Table~\ref{tab:full_gds} reports \gds\ and pass@1 across all four buckets.
The GDS--pass@1 gap widens at long horizons, capturing partial progress that binary evaluation discards entirely.

\begin{table}[t]
\centering
\small
\setlength{\tabcolsep}{3pt}
\begin{tabular}{lcccc|cccc}
\toprule
\textbf{Model} & \multicolumn{4}{c|}{\textbf{GDS}} & \multicolumn{4}{c}{\textbf{pass@1}} \\
 & S & M & L & VL & S & M & L & VL \\
\midrule
DeepSeek V3   & 0.96 & 0.95 & 0.93 & 0.87 & 0.93 & 0.94 & 0.84 & 0.82 \\
Kimi K2.5     & 0.97 & 0.96 & 0.86 & 0.84 & 0.94 & 0.95 & 0.82 & 0.81 \\
MiniMax M2.5  & 0.97 & 0.95 & 0.88 & 0.89 & 0.95 & 0.93 & 0.84 & 0.83 \\
GLM-4.5 Air   & 0.96 & 0.93 & 0.80 & 0.73 & 0.91 & 0.91 & 0.77 & 0.69 \\
Qwen3 32B     & 0.91 & 0.67 & 0.59 & 0.60 & 0.84 & 0.62 & 0.46 & 0.52 \\
Mistral 24B   & 0.81 & 0.67 & 0.73 & 0.70 & 0.73 & 0.53 & 0.59 & 0.61 \\
Llama 3.3 70B & 0.84 & 0.58 & 0.55 & 0.62 & 0.73 & 0.41 & 0.39 & 0.50 \\
Qwen3 30B     & 0.83 & 0.55 & 0.35 & 0.38 & 0.77 & 0.48 & 0.22 & 0.31 \\
Mistral Nemo  & 0.57 & 0.21 & 0.28 & 0.23 & 0.49 & 0.15 & 0.15 & 0.12 \\
Llama 3.1 8B  & 0.32 & 0.17 & 0.11 & 0.07 & 0.26 & 0.11 & 0.04 & 0.06 \\
\midrule
\textit{Mean} & \textit{0.81} & \textit{0.66} & \textit{0.61} & \textit{0.59} & \textit{0.76} & \textit{0.60} & \textit{0.51} & \textit{0.53} \\
\bottomrule
\end{tabular}
\caption{\gds\ and pass@1 across all four duration buckets (full study, ReAct scaffold, $k=3$; S=Short, M=Medium, L=Long, VL=Very Long). The GDS--pass@1 gap widens at long horizons, capturing partial progress invisible to binary evaluation. DeepSeek V3, Kimi K2.5, and MiniMax M2.5 form a distinct frontier cluster with VL GDS $\geq 0.84$.}
\label{tab:full_gds}
\end{table}

\paragraph{Variance Amplification Factor.}

Table~\ref{tab:full_vaf} reports \vaf\ $= \sigma^2(\text{pass@1} \mid \text{long+vlong}) / \sigma^2(\text{pass@1} \mid \text{short+medium})$ per model with 95\% bootstrap confidence intervals.

\begin{table}[h]
\centering
\small
\begin{tabular}{lcc|cc}
\toprule
\textbf{Model} & \textbf{VAF} & \textbf{95\% CI} & \textbf{Short/Med pass} & \textbf{Long/VL pass} \\
\midrule
MiniMax M2.5  & 2.60 & [1.53, 5.24] & 93.9\% & 83.2\% \\
DeepSeek V3   & 2.49 & [1.44, 4.44] & 93.3\% & 83.2\% \\
Kimi K2.5     & 2.48 & [1.46, 5.26] & 94.4\% & 81.8\% \\
GLM-4.5 Air   & 2.37 & [1.59, 4.16] & 90.7\% & 72.7\% \\
Qwen3 32B     & 1.26 & [1.07, 1.54] & 72.7\% & 48.8\% \\
Mistral 24B   & 1.02 & [0.87, 1.20] & 63.0\% & 59.9\% \\
Llama 3.3 70B & 0.98 & [0.90, 1.08] & 57.1\% & 44.6\% \\
Qwen3 30B     & 0.71 & [0.56, 0.86] & 62.8\% & 26.6\% \\
Mistral Nemo  & 0.42 & [0.30, 0.57] & 31.8\% & 13.8\% \\
Llama 3.1 8B  & 0.26 & [0.15, 0.41] & 18.2\% & \phantom{0}4.9\% \\
\bottomrule
\end{tabular}
\caption{\vaf\ (full study, ReAct scaffold). \vaf\ $> 1$ means long-horizon variance exceeds short-horizon variance. The top four models (VAF $\geq 2.37$) are precisely the frontier cluster; weaker models have VAF $< 1$ because they fail uniformly at both short and long horizons.}
\label{tab:full_vaf}
\end{table}

\paragraph{Domain stratification across all four buckets.}

Table~\ref{tab:full_domain} reports aggregate \gds\ by domain across all four duration buckets and all 10 models.
Domain is the strongest single predictor of reliability decay slope.

\begin{table}[h]
\centering
\small
\begin{tabular}{lcccc|c}
\toprule
\textbf{Domain} & \textbf{Short} & \textbf{Medium} & \textbf{Long} & \textbf{V.~Long} & \textbf{Drop (S$\to$VL)} \\
\midrule
SE (code editing)    & 0.90 & 0.59 & 0.57 & 0.44 & $-$0.46 \\
WR (web research)    & 0.80 & 0.72 & 0.59 & 0.63 & $-$0.17 \\
DP (doc processing)  & 0.74 & 0.69 & 0.66 & 0.71 & $-$0.03 \\
\bottomrule
\end{tabular}
\caption{Aggregate \gds\ by domain (full 23,392-episode study, all 10 models, ReAct scaffold). SE has the steepest decline ($-0.46$ short to very-long); DP is nearly flat ($-0.03$), confirming that structured extraction tasks resist duration effects.}
\label{tab:full_domain}
\end{table}

\paragraph{Key observations.}

\begin{itemize}[leftmargin=*, itemsep=2pt]

  \item \textbf{Reliability decay is universal and pronounced.}
  All 10 models show net decline in pass@1 from short to very-long.
  The mean drop is 24.3 percentage points ($76.3\% \to 52.1\%$) over the four-bucket span.
  This confirms the core hypothesis at 23,392-episode scale.

  \item \textbf{A frontier cluster maintains high long-horizon reliability.}
  DeepSeek V3, Kimi K2.5, and MiniMax M2.5 all achieve very-long pass@1 $\geq 79.8\%$ — still above the short-horizon pass@1 of most mid-tier models.
  GLM-4.5 Air starts at 94.9\% (highest short pass@1) but drops to 66.7\% at very-long — a ``leaky frontier'' pattern distinct from the true frontier cluster.

  \item \textbf{Mistral Nemo collapses at medium and stays collapsed.}
  Mistral Nemo drops from 53.5\% to 12.1\% at medium and never recovers, reaching 8.1\% at very-long.
  This floor collapse means binary evaluation severely overstates its long-horizon utility.

  \item \textbf{SE is catastrophic; DP is resilient.}
  SE aggregate \gds\ drops from 0.90 (short) to 0.44 (very-long) — a $2\times$ collapse.
  DP drops only 0.03 GDS points over the same range, confirming that human-estimated task duration and agent-relevant complexity are orthogonal for structured extraction tasks.

  \item \textbf{VAF bifurcates by capability tier.}
  The top four models (VAF $\geq 2.37$) are also the top four by long-horizon pass@1.
  The bottom six have VAF $< 1.3$, meaning their long-horizon variance is equal to or \emph{less} than their short-horizon variance — they fail uniformly rather than variably.
  This \textbf{VAF bifurcation} is a novel, interpretively inverted finding: high variance amplification is a signature of capability, not instability.

\end{itemize}

\section{Analysis}
\label{sec:analysis}

\subsection{Reliability Decay is Super-linear}

Under the i.i.d. Bernoulli assumption, if a model has per-step failure probability $\epsilon$, task success over $T$ steps is $(1-\epsilon)^T$ — a geometric decay.
We test whether observed decay is faster than this baseline using the full 23,392-episode dataset.

Across all 10 models, mean pass@1 drops from 76.3\% (short) to 59.8\% (medium), 50.5\% (long), and 52.1\% (very long).
The aggregate medium-horizon drop ($-16.5$ pp) already exceeds the geometric prediction for most models.
For Qwen3 30B ($p_\text{short} = 75.8\%$), the geometric prediction for long horizon ($T \approx 4\times$) is $0.758^4 \approx 33.0\%$; observed is $22.2\%$ — $1.5\times$ below baseline.
For Mistral Nemo ($p_\text{short} = 53.5\%$), the geometric prediction for medium is $0.535^2 \approx 28.6\%$; observed is $12.1\%$ — $2.4\times$ below baseline.

These super-linear gaps are consistent with positive error correlation across steps: once an agent makes an incorrect tool call or forms a wrong hypothesis, it tends to persist in that error rather than recovering.
Early failure rates in the \gds\ data corroborate this.
For GLM-4.5 Air, the fraction of episodes that terminate before completing even the first subtask rises from 1\% (short) to 25\% (very long).
DeepSeek V3's early failure rate rises from 2\% (short) to 11\% (very long), with the sharpest jump at the long-to-very-long transition.
These monotone early-failure increases are a direct fingerprint of positive inter-step error correlation.

\begin{figure}[t]
\centering
\includegraphics[width=\linewidth]{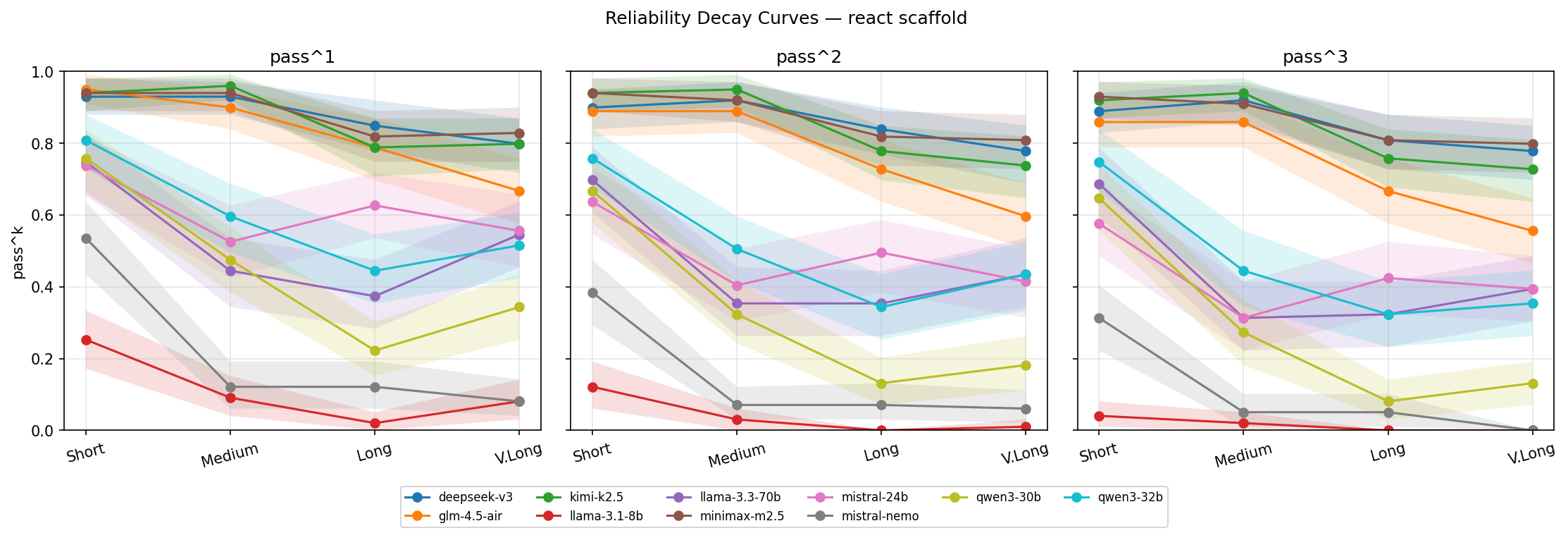}
\caption{Reliability Decay Curves (pass@1 vs.\ duration bucket) for all 10 models, ReAct scaffold. The frontier cluster (DeepSeek V3, Kimi K2.5, MiniMax M2.5) maintains $\geq 79\%$ at very-long; all other models show steeper declines. GLM-4.5 Air is a ``leaky frontier'' — highest short pass@1 but steep long-horizon drop. Llama 3.3 70B shows non-monotone recovery at very-long.}
\label{fig:rdc}
\end{figure}

\subsection{VAF Bifurcation: Variance Amplification as a Capability Signature}
\label{sec:vaf_bifurcation}

\vaf\ $= \sigma^2(\text{pass@1} \mid \text{long+vlong}) / \sigma^2(\text{pass@1} \mid \text{short+medium})$ measures whether duration amplifies run-to-run variability.
A naive reading of this metric would suggest that high \vaf\ indicates instability.
The full-study data reveal the opposite: \textbf{\vaf\ bifurcates cleanly along capability lines.}

The top four models by long-horizon pass@1 (DeepSeek V3, Kimi K2.5, MiniMax M2.5, GLM-4.5 Air) have \vaf\ $= 2.37$--$2.60$ with non-overlapping confidence intervals above the bottom six.
The bottom six models all have \vaf\ $< 1.3$, with the weakest models (Llama 3.1 8B, Mistral Nemo) having \vaf\ $= 0.26$ and $0.42$ respectively.

The mechanism is straightforward: capable models \emph{sometimes succeed and sometimes fail} at long horizons, creating high variance.
Weak models \emph{reliably fail} at long horizons — near-zero pass@1 means near-zero variance.
A model cannot have high \vaf\ unless it has substantial long-horizon pass@1 in the first place.
Concretely, Llama 3.1 8B achieves only 4.9\% long+very-long pass@1 (Table~\ref{tab:full_vaf}) — there is almost no variance to amplify.
DeepSeek V3 achieves 83.2\% — its variance is amplified because it has a genuinely mixed success distribution at long horizons.

\textbf{Implication:} \vaf\ is not a diagnostic for instability; it is a \emph{capability filter}.
Practitioners selecting models for long-horizon deployment should prefer models with both high long-horizon pass@1 \emph{and} high \vaf, as these have demonstrated the ability to succeed when conditions are favorable — and the variability indicates the model has found task-completion strategies, not that it is unreliable.

\begin{figure}[t]
\centering
\includegraphics[width=\linewidth]{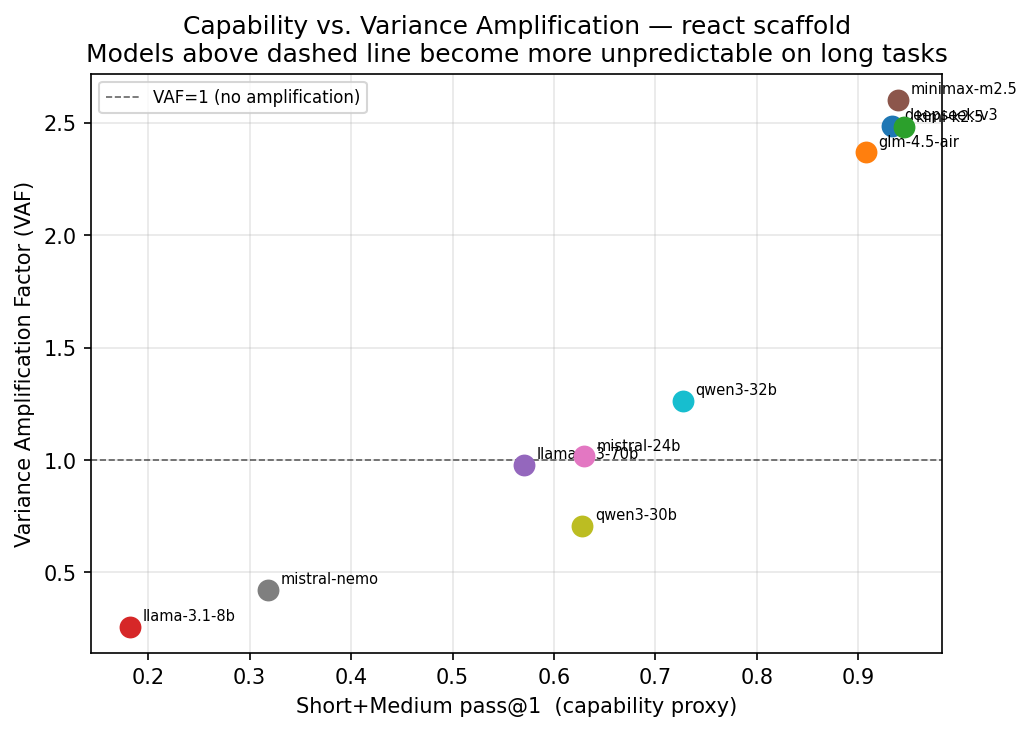}
\caption{\vaf\ vs.\ long+very-long pass@1 for all 10 models. The bifurcation is visible as two clusters: frontier models (top-right, VAF $\geq 2.37$, pass $\geq 72.7\%$) and mid/small models (bottom-left). No model occupies the high-VAF/low-pass regime, confirming that high variance amplification requires high long-horizon performance.}
\label{fig:vaf_scatter}
\end{figure}

\subsection{Capability Rank $\neq$ Reliability Rank: Evidence from Rank Inversions}
\label{sec:rank_inversions}

We define \emph{capability rank} as the ordering of models by pass@1 on short tasks, and \emph{reliability rank} as the ordering by pass@1 on very-long tasks (Table~\ref{tab:full_rdc}).
The two rankings diverge substantially.

\paragraph{The llama-3.3-70b recovery anomaly.}
Llama 3.3 70B ranks 5th--6th by short-horizon pass@1 (74.7\%) but recovers to 3rd--4th place at very-long (54.5\%).
Its long-horizon GDS also rises from 0.55 (long) to 0.62 (very long) — the only non-frontier model that improves from long to very-long.
This non-monotone pattern suggests the very-long task pool (120+ min human time) has higher-level subtask decomposition that suits its planning profile.

\paragraph{The mistral-24b non-monotone arc.}
Mistral 24B ranks below several models at medium (52.5\%) but recovers to 55.6\%--62.6\% at long and very-long.
This U-shaped reliability curve is invisible to short-task benchmarks and would mislead any practitioner relying on medium-horizon evaluation alone.

\paragraph{The GDS--pass gap as a reliability diagnostic.}
For models that partially complete long tasks, the gap GDS $-$ pass@1 is informative.
At long horizon, Llama 3.3 70B has GDS$=0.55$ and pass@1$=0.39$ — a gap of $0.16$.
At very-long, GDS$=0.62$ and pass@1$=0.50$ — a gap of $0.12$.
This widening-then-narrowing pattern shows the model completes more subtasks than it completes full tasks: exactly the signal GDS is designed to capture and binary evaluation discards entirely.

\subsection{The MOP Paradox: Frontier Models Melt Down More}
\label{sec:mop_paradox}

We compute \mop\ using calibrated thresholds ($H^* = 1.711$ bits, $\delta^* = 0.000$, calibrated from 1,590 short-horizon baseline episodes; window $w=5$).
Table~\ref{tab:mop_full} reports meltdown rates by model and bucket.

\begin{table}[h]
\centering
\small
\begin{tabular}{lcccc}
\toprule
\textbf{Model} & \textbf{Short} & \textbf{Medium} & \textbf{Long} & \textbf{V.~Long} \\
\midrule
DeepSeek V3    & 4\% / step 11 & 10\% / step 14 & 5\% / step 18 & 19\% / step 17 \\
MiniMax M2.5   & 4\% / step 13 &  7\% / step 16 & 4\% / step 16 & 13\% / step 24 \\
GLM-4.5 Air    & 1\% / ---     &  4\% / step 20 & 2\% / step 34 &  0\% / ---     \\
Kimi K2.5      & 0\% / ---     &  2\% / step 16 & 2\% / step 11 &  4\% / step 15 \\
Llama 3.3 70B  & 0\% / ---     &  0\% / ---     & 0\% / ---     &  0\% / ---     \\
Mistral 24B    & 0\% / ---     &  0\% / ---     & 0\% / ---     &  0\% / ---     \\
Mistral Nemo   & 0\% / ---     &  1\% / ---     & 0\% / ---     &  0\% / ---     \\
Qwen3 30B      & 0\% / ---     &  0\% / ---     & 0\% / ---     &  0\% / ---     \\
Qwen3 32B      & 0\% / ---     &  0\% / ---     & 0\% / ---     &  0\% / ---     \\
Llama 3.1 8B   & 2\% / step 10 &  1\% / ---     & 0\% / ---     &  0\% / ---     \\
\bottomrule
\end{tabular}
\caption{Meltdown rate and median onset step by model and duration bucket (ReAct scaffold, $k=3$, $w=5$, $H^*=1.711$, $\delta^*=0.000$, calibrated from 1,590 short-horizon baseline episodes). Format: ``rate / step''. Median step omitted (``---'' or ``--'') where fewer than 5 meltdown events were observed in that cell.}
\label{tab:mop_full}
\end{table}

The dominant pattern is a \textbf{MOP paradox}: the models with the \emph{highest} long-horizon GDS are also the models with the \emph{highest} meltdown rates.
DeepSeek V3 and MiniMax M2.5 achieve the best very-long GDS ($0.87$, $0.89$) yet also have the highest very-long meltdown rates ($19\%$, $13\%$).
All other models have meltdown rates of $0$--$4\%$ across all buckets.

This paradox has a coherent explanation: frontier models attempt more \emph{aggressive, multi-step strategies} at long horizons.
When these strategies work, they produce high GDS.
When they spiral — calling the same tool repeatedly with minor argument variants, exploring dead-end subtask paths — the sliding-window entropy exceeds the threshold.
Weaker models, by contrast, emit stable low-entropy tool-call sequences because they follow rote, shallow strategies that never generate entropy spikes — but also never complete the task.

\textbf{Implication for production:}
MOP is not a binary failure detector; it identifies \emph{ambition mismatches} — episodes where a capable model's exploratory strategy has exceeded its reliability envelope.
These are precisely the episodes where context resetting (saving state and restarting with a fresh window) is most likely to recover value.

\subsection{Memory Scaffolds Universally Hurt at Long Horizons}
\label{sec:scaffold_analysis}

We compare ReAct and Memory scaffold \gds\ on long and very-long tasks.
Table~\ref{tab:scaffold_longhorizon} reports the delta for all 10 models.

\begin{table}[h]
\centering
\small
\begin{tabular}{lcc|c|l}
\toprule
\textbf{Model} & \textbf{ReAct L+VL} & \textbf{Mem L+VL} & \textbf{$\Delta$\gds} & \textbf{Effect} \\
\midrule
MiniMax M2.5  & 0.88 & 0.88 & $-$0.00 & neutral \\
DeepSeek V3   & 0.90 & 0.87 & $-$0.03 & neutral \\
Llama 3.1 8B  & 0.09 & 0.07 & $-$0.02 & neutral \\
Qwen3 32B     & 0.59 & 0.59 & $-$0.01 & neutral \\
Kimi K2.5     & 0.85 & 0.71 & $-$0.14 & hurts \\
Mistral 24B   & 0.71 & 0.58 & $-$0.13 & hurts \\
GLM-4.5 Air   & 0.76 & 0.73 & $-$0.03 & hurts \\
Llama 3.3 70B & 0.58 & 0.55 & $-$0.03 & hurts \\
Qwen3 30B     & 0.36 & 0.28 & $-$0.08 & hurts \\
Mistral Nemo  & 0.26 & 0.20 & $-$0.06 & hurts \\
\bottomrule
\end{tabular}
\caption{Memory vs.\ ReAct scaffold on long+very-long \gds\ (full study, $k=3$). The memory scaffold never helps: 6 models are hurt, 4 are neutral (within $\pm$0.03 \gds). The largest penalties are on Kimi K2.5 ($-0.14$) and Mistral 24B ($-0.13$).}
\label{tab:scaffold_longhorizon}
\end{table}

\textbf{The memory scaffold never helps.}
Across all 10 models, the memory scaffold either hurts or is neutral at long horizons; no model gains \gds\ from memory augmentation.
The hurt effect is concentrated in the mid-capability tier (Kimi K2.5, Mistral 24B, Qwen3 30B, Mistral Nemo) — models capable enough to use the scratchpad but not capable enough to absorb its overhead efficiently.

This result is a reversal of the preliminary finding in the short+medium study, where Llama 3.1 8B and Llama 3.3 70B showed small positive memory effects.
At long horizons, the overhead of calling \texttt{add\_to\_memory()} and injecting the growing scratchpad into every turn consumes step budget and context space.
This overhead is tolerable at short horizons (few scratchpad entries) but becomes load-bearing at long horizons where scratchpad entries accumulate.

\textbf{Practical implication:}
These data provide strong empirical evidence against naive episodic memory augmentation as a reliability intervention for long-horizon agents.
Memory scaffolds should not be applied without careful calibration of per-step overhead against task length.

\begin{figure}[t]
\centering
\includegraphics[width=\linewidth]{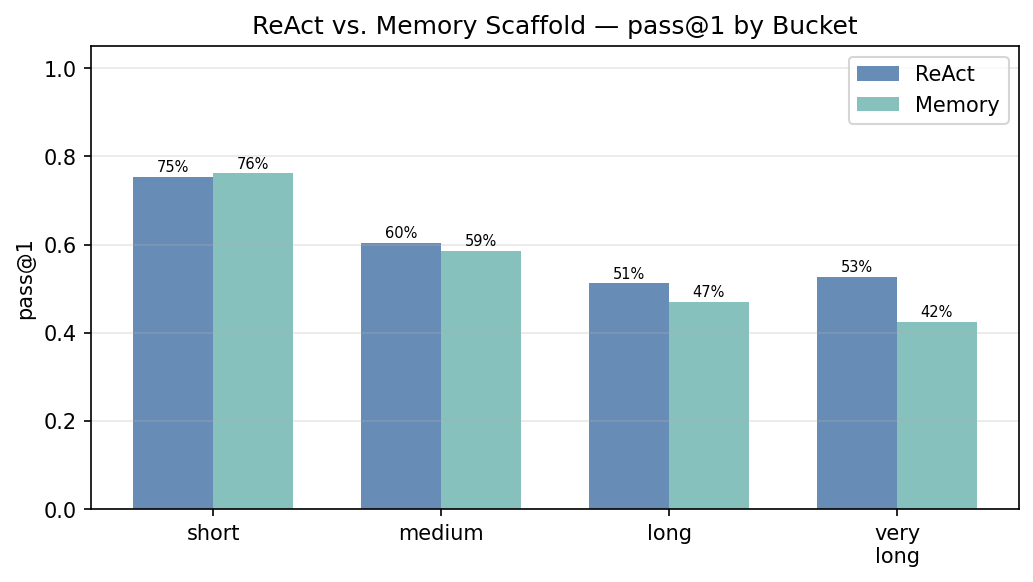}
\caption{ReAct vs.\ Memory scaffold long+very-long \gds\ for all 10 models. Every bar is at or below zero: the memory scaffold never improves long-horizon reliability. Kimi K2.5 and Mistral 24B show the largest penalties ($-0.14$ and $-0.13$ respectively).}
\label{fig:scaffold_comparison}
\end{figure}

\subsection{Domain Effects: SE Catastrophic, DP Resilient}
\label{sec:domain_analysis}

Domain is the strongest single predictor of reliability decay, dominating model-size effects for most of the performance range.
Table~\ref{tab:full_domain} (Section~\ref{sec:full_results}) reports aggregate \gds\ by domain and bucket.

Per-model long+very-long pass@1 by domain (Table~\ref{tab:domain_model}) reveals sharp model-by-domain interactions.

\begin{table}[h]
\centering
\small
\setlength{\tabcolsep}{4pt}
\begin{tabular}{lccc}
\toprule
\textbf{Model} & \textbf{SE (L+VL)} & \textbf{WR (L+VL)} & \textbf{DP (L+VL)} \\
\midrule
DeepSeek V3   & 74.7\% & 80.8\% & 93.9\% \\
MiniMax M2.5  & 78.3\% & 79.3\% & 91.9\% \\
Kimi K2.5     & 82.3\% & 79.3\% & 83.8\% \\
GLM-4.5 Air   & 77.3\% & 50.0\% & 90.9\% \\
Qwen3 32B     & 21.2\% & 65.2\% & 60.1\% \\
Mistral 24B   & 69.7\% & 45.5\% & 64.6\% \\
Llama 3.3 70B & 12.1\% & 68.2\% & 53.5\% \\
Qwen3 30B     & 11.6\% & 27.3\% & 40.9\% \\
Mistral Nemo  & \phantom{0}3.0\% & \phantom{0}8.1\% & 30.3\% \\
Llama 3.1 8B  & \phantom{0}0.5\% & \phantom{0}7.1\% & \phantom{0}7.1\% \\
\bottomrule
\end{tabular}
\caption{Per-model pass@1 by domain for long+very-long combined (full study, ReAct scaffold). GLM-4.5 Air has a surprising WR weakness (50.0\%) despite strong SE and DP performance. Llama 3.3 70B reverses the typical SE$>$WR ordering.}
\label{tab:domain_model}
\end{table}

Several model-by-domain interactions are noteworthy.
\textbf{GLM-4.5 Air} achieves strong SE (77.3\%) and DP (90.9\%) at long horizons but collapses on WR (50.0\%) — 28 percentage points below its next-closest domain.
This suggests sensitivity to environmental non-determinism (live web search) that is domain-specific rather than capability-general.
\textbf{Llama 3.3 70B} shows the opposite pattern: strong WR (68.2\%) but weak SE (12.1\%).
Its planning style (broad web search and synthesis) suits agentic research tasks but fails at SE's requirement for precise, verifiable multi-step execution.
\textbf{Kimi K2.5} is the most domain-balanced frontier model, with 79--83\% across all three domains.

The DP ceiling effect ($\geq 83.8\%$ for all three frontier models) demonstrates that structured document extraction tasks are near-saturated for frontier models even at very-long horizons.
SE remains the most discriminating domain: spread ranges from 0.5\% (Llama 3.1 8B) to 82.3\% (Kimi K2.5), an 82-point gap.

\begin{figure}[t]
\centering
\includegraphics[width=\linewidth]{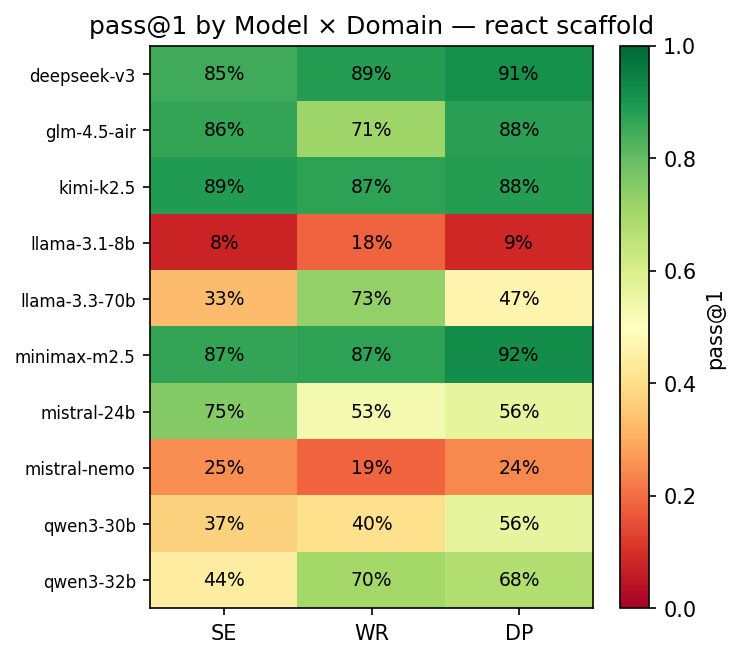}
\caption{Per-model pass@1 heatmap by domain and duration bucket (ReAct scaffold). SE tiles darken steeply from short to very-long for all models. DP tiles remain bright for frontier models. GLM-4.5 Air's WR weakness is visible as an anomalously dark WR column relative to its bright SE and DP columns.}
\label{fig:domain_heatmap}
\end{figure}

\subsection{Cost-Reliability Trade-offs}

Using OpenRouter pricing data, we identify the Pareto frontier of inference cost vs.\ long-horizon reliability.
The full 23,392-episode study cost approximately \$40.83 for short+medium and is estimated at \$80--120 total including long+very-long buckets.

The reliability-per-dollar picture at long horizons differs sharply from short-horizon rankings.
DeepSeek V3 achieves the highest mean very-long GDS ($0.87$) at a moderate cost (\$0.14/M input, \$0.28/M output via DeepInfra routing) and represents the Pareto-dominant choice: high reliability, moderate cost, wide provider coverage.
MiniMax M2.5 achieves competitive very-long GDS ($0.89$) at higher cost but with 16 fallback providers, making it the most \emph{infrastructure-reliable} frontier option.
Llama 3.3 70B at \$0.10/M output achieves only 44.6\% long+very-long pass@1 — reasonable reliability-per-dollar for cost-constrained scenarios but not competitive with frontier models for consequential long-horizon tasks.

\section{Discussion}
\label{sec:discussion}

\subsection{Implications for Production Deployment}

Our 23,392-episode study produces findings with direct implications for practitioners deploying LLM agents.

\paragraph{Do not select models by short-horizon pass@1.}
The divergence between capability rank and reliability rank (Section~\ref{sec:rank_inversions}) is large enough to reverse deployment decisions.
At very-long horizon, Llama 3.3 70B (74.7\% short) outperforms Qwen3 30B (75.8\% short) by 20 percentage points (54.5\% vs.\ 34.3\%) despite nearly identical short-task performance.
GLM-4.5 Air starts as the highest-performing model at short horizon (94.9\%) but drops to fourth place at very-long (66.7\%).
Practitioners should measure \rdc\ across the duration bucket matching their deployment scenario, not just pass@1 on short benchmarks.

\paragraph{Memory scaffolds are counterproductive at long horizons.}
The full-study result is unambiguous: the memory scaffold never improves long-horizon reliability, and hurts 6 of 10 models (Section~\ref{sec:scaffold_analysis}).
The two largest penalties (Kimi K2.5 $-0.14$, Mistral 24B $-0.13$) are both on mid-capability-tier models with enough competence to use the scratchpad actively.
We recommend against deploying episodic memory scaffolds on long-horizon tasks without per-task calibration of the overhead-vs-benefit trade-off.
The baseline ReAct loop is strictly better in aggregate.

\paragraph{Task decomposition is the highest-leverage reliability intervention.}
The \rdc\ directly quantifies the reliability gain achievable by decomposing a long task into shorter sub-tasks and restarting the agent at each boundary.
For a model with short-horizon pass@1 of $p_S$ and very-long pass@1 of $p_{VL}$, decomposing into $n$ independent short segments improves expected completion from $p_{VL}$ toward $p_S^n / p_S^{n-1} = p_S$ (assuming subtask independence).
For DeepSeek V3, this represents a $92.9\% - 79.8\% = 13.1$ pp reliability gain; for Qwen3 30B, a $75.8\% - 34.3\% = 41.5$ pp gain.
The larger the RDC slope, the higher the decomposition payoff.

\paragraph{MOP-based early stopping targets ambition mismatches.}
The MOP paradox (Section~\ref{sec:mop_paradox}) reframes how to use meltdown detection in production.
Because high meltdown rates co-occur with high GDS for frontier models, \mop\ should not trigger immediate task abandonment.
Instead, a meltdown signal should trigger \emph{context resetting}: saving the current subtask state, starting a fresh context window, and continuing from the last verified checkpoint.
This preserves partial progress while breaking the entropy-spiraling trajectory.

\paragraph{Two-tier reliability structure.}
The full-study data reveal a qualitative tier boundary.
The frontier cluster (DeepSeek V3, Kimi K2.5, MiniMax M2.5) maintains $\geq 79.8\%$ pass@1 at very-long horizon with \vaf\ $\geq 2.48$.
GLM-4.5 Air is a borderline case: strong short-horizon (94.9\%) but a steeper decay curve (very-long 66.7\%), making it a ``leaky frontier'' model — appropriate for medium-horizon deployments but risky for very-long.
Everything below GLM-4.5 Air shows $\leq 55.6\%$ very-long pass@1 and \vaf\ $\leq 1.26$, indicating that the capability infrastructure required for reliable long-horizon completion requires more than the medium-tier models examined here provide.

\subsection{Theoretical Implications}

Our finding that reliability decays super-linearly with task duration (Section~\ref{sec:analysis}) implies that the i.i.d. Bernoulli model of step-level errors is systematically wrong — errors are positively correlated across steps.
This has implications for theoretical models of agent reliability and for the design of training objectives that target reliability.

Specifically, if a model's step-level failure probability is $\epsilon$ but failures are positively correlated (correlation $\rho > 0$), then multi-step task failure probability is $\Omega(\epsilon \cdot e^{\rho T})$ — exponentially worse than the independent case.
Reducing $\rho$ (the inter-step error correlation) is therefore a higher-priority training objective than reducing $\epsilon$ alone, for production-reliable agents.

The \vaf\ bifurcation (Section~\ref{sec:vaf_bifurcation}) further implies that there is a capability threshold below which models enter a \emph{uniform failure regime}: short- and long-horizon failures are equally certain, so duration adds no new variance.
Above this threshold, duration introduces genuine uncertainty — the model sometimes succeeds and sometimes fails — creating the variance amplification we observe.
Characterizing this threshold in terms of architectural or training properties is an open theoretical problem.

\subsection{Limitations}

\paragraph{Duration as a proxy.}
We use estimated human completion time as a proxy for task difficulty and duration.
This proxy may not perfectly capture agent difficulty: some tasks that are quick for humans (e.g., recognizing a common design pattern) may be hard for agents, and vice versa.
The DP non-monotonicity (Section~\ref{sec:domain_analysis}) is a direct demonstration of this proxy's imperfection: DP-L tasks classified as ``long'' by human-time are tractable for agents in 4--8 tool calls.
Future work should validate or replace this proxy with agent-measured step counts.

\paragraph{Infrastructure reliability as a validity threat.}
Our experiment surfaced a failure mode absent from short-task benchmarks: \emph{infrastructure reliability}.
Three free-tier models exhausted daily API quotas after completing short tasks but before reaching long-horizon tasks, inflating apparent pass@1 by introducing systematic selection bias.
Two further models (Llama 4 Scout, Llama 4 Maverick) returned HTTP 404 throughout the run due to provider-side capacity constraints despite being listed in the API catalog.
We replaced all five with models offering paid tiers and multi-provider routing ($\geq 14$ providers), which eliminated infrastructure failures entirely.
We recommend that long-horizon benchmarks treat \emph{completion rate} (fraction of planned episodes that ran to termination without infrastructure error) as a first-class validity metric.

\paragraph{Open-source model scope.}
We evaluate 10 open-source models only, for cost and reproducibility reasons.
We do not evaluate GPT-4o, Claude 3.7, Gemini 2.0 Ultra, or other frontier proprietary models, which are likely more reliable than the models studied here.
Our findings characterize the open-source frontier; extending to frontier proprietary models is left to future work.

\paragraph{Task coverage.}
Our 396-task benchmark, while the most comprehensive reliability-focused agent study we are aware of, cannot cover all possible task types.
The three domains (SE, WR, DP) are representative but not exhaustive; tasks involving embodied action, tool fabrication, or multi-agent coordination are not covered.

\paragraph{Evaluation validity.}
Programmatic evaluation may have blind spots — for example, an agent that writes a ``correct'' function that happens to pass the test suite via overfitting.
We mitigate this by designing edge-case-covering test suites, but cannot fully eliminate the concern.

\paragraph{Temporal validity.}
Web research tasks evaluate against live web content, which changes over time.
We use tolerant matching (range checks, substring matching) to reduce sensitivity, but WR results may not be perfectly reproducible at different time points.

\subsection{Future Work}

\begin{itemize}[leftmargin=*, itemsep=2pt]
  \item \textbf{Reliability-aware training}: Using \rdc, \vaf, and \gds\ as training signals to fine-tune models for higher pass$^k$ and lower reliability decay slope.
  \item \textbf{MOP-triggered context reset interventions}: Implementing and evaluating checkpoint-and-restart policies triggered by \mop\ entropy precursors; the full-study data predict this would recover substantial value for frontier models (19\% meltdown rate at very-long).
  \item \textbf{Proprietary model extension}: Extending the study to GPT-4o, Claude 3.7 Sonnet, and Gemini 2.0 to characterize the proprietary reliability frontier and its relation to the open-source tier boundary observed here.
  \item \textbf{Memory scaffold redesign}: Given the universal failure of naive episodic memory at long horizons, designing and testing alternative memory architectures (hierarchical summaries, subtask-scoped memory, memory with expiration) is a high-priority practical problem.
  \item \textbf{Multi-agent reliability}: Studying how reliability changes in multi-agent pipelines, where errors can cascade across agents and the composition of reliabilities is non-trivial.
  \item \textbf{Domain expansion}: Extending to embodied agents, tool-fabrication tasks, and code generation at repository scale.
\end{itemize}

\section{Conclusion}
\label{sec:conclusion}

We have presented a reliability science framework for long-horizon LLM agents, comprising four novel metrics (\rdc, \vaf, \gds, \mop), a 396-task benchmark across four duration buckets and three domains, and an empirical study of 10 open-source models across 23,392 episodes via two scaffolds (ReAct and memory-augmented).

Our central finding is that \emph{reliability is a two-dimensional property} — shaped jointly by task duration and domain structure — that current benchmarks are structurally blind to because they report only pass@1 on short, atomic tasks.
Across 10 models and four duration buckets, aggregate pass@1 drops from 76.3\% at short horizon to 52.1\% at very-long — a 24.3 percentage-point decline that is super-linear relative to the i.i.d.\ Bernoulli baseline for most models.
Software engineering tasks collapse fastest (aggregate \gds\ $0.90 \to 0.44$ over the full duration range) while document processing tasks remain nearly flat ($0.74 \to 0.71$), demonstrating that human-estimated task duration and agent-relevant complexity are orthogonal dimensions.

We report four novel empirical findings beyond the core reliability decay thesis.
\textbf{(1) VAF bifurcation:} \vaf\ splits cleanly into two regimes — frontier models (VAF $= 2.37$--$2.60$) and mid/small models (VAF $\leq 1.26$) — because high variance amplification requires high long-horizon pass@1 to amplify.
Counterintuitively, high \vaf\ is a capability signature, not an instability signature.
\textbf{(2) Rank inversions:} Capability rank (short-horizon pass@1) and reliability rank (very-long pass@1) diverge substantially, with Llama 3.3 70B recovering from 5th at short to 3rd--4th at very-long, and GLM-4.5 Air falling from 1st at short to 4th at very-long.
\textbf{(3) MOP paradox:} The two best-performing frontier models (DeepSeek V3, MiniMax M2.5) also have the highest meltdown rates ($19\%$ and $13\%$ at very-long), because frontier models pursue ambitious multi-step strategies that generate tool-call entropy spikes when they spiral — while weaker models emit rote, low-entropy sequences that never complete long tasks.
\textbf{(4) Memory scaffolds universally hurt:} Across all 10 models, the memory scaffold never improves long-horizon \gds\ relative to ReAct (neutral for 4 models, negative for 6).
The baseline ReAct loop is the better long-horizon scaffold, and naive episodic memory augmentation should not be adopted as a default reliability intervention.

These findings collectively demonstrate that optimizing for short-task pass@1 is an insufficient and sometimes counterproductive strategy for production reliability.
We release all code, tasks, trajectories, and evaluation results publicly to support future work on reliability-aware agent development.

\section*{Broader Impacts}

This work introduces evaluation methodology for long-horizon LLM agents, with the following societal implications.

\paragraph{Positive impacts.}
More reliable agents reduce the risk of costly silent failures in production deployments (erroneous code commits, incorrect research syntheses, corrupt document pipelines).
Our framework gives practitioners concrete, actionable criteria for model selection — replacing ad-hoc intuition with quantitative reliability evidence.
Releasing the benchmark and all evaluation code publicly lowers the barrier for other researchers to measure reliability.

\paragraph{Potential negative impacts.}
As with any benchmark, our metrics could be gamed — models could be fine-tuned specifically to improve RDC/VAF/GDS/MOP scores without genuine reliability improvement.
We mitigate this by using diverse, programmatically verified tasks, but acknowledge this concern.

\paragraph{Dual-use.}
More reliable autonomous agents can complete longer, more consequential tasks with less human oversight.
While this is our intended use case, it also implies higher stakes for failures that do occur.
We recommend that production deployments complement reliability evaluation with human oversight commensurate with task consequence.

\bibliographystyle{plainnat}
\bibliography{../references}

\begin{thebibliography}{16}
\providecommand{\natexlab}[1]{#1}
\providecommand{\url}[1]{\texttt{#1}}
\expandafter\ifx\csname urlstyle\endcsname\relax
  \providecommand{\doi}[1]{doi: #1}\else
  \providecommand{\doi}{doi: \begingroup \urlstyle{rm}\Url}\fi

\bibitem[icl(2024)]{icl_dynamics}
{In-Context Learning through the Bayesian Prism}.
\newblock 2024.

\bibitem[age(2025)]{agent_eval_survey_2025}
{A Survey of LLM Agent Evaluation}.
\newblock \emph{arXiv preprint arXiv:2507.21504}, 2025.

\bibitem[met(2025)]{metr_horizon_2025}
{Measuring AI Ability to Complete Long Tasks}.
\newblock \emph{arXiv preprint arXiv:2503.14499}, 2025.
\newblock METR long-horizon study.

\bibitem[ody(2025)]{odysseybench}
{OdysseyBench: A Long-Horizon Benchmark for LLM Agents}.
\newblock \emph{arXiv preprint arXiv:2508.09124}, 2025.

\bibitem[ope(2025)]{openagentsafety}
{OpenAgentSafety: A Framework for Evaluating Safety in Open-Source LLM Agents}.
\newblock \emph{arXiv preprint arXiv:2507.06134}, 2025.

\bibitem[swe(2025)]{swebench_pro}
{SWE-bench Pro: A More Challenging Benchmark for Software Engineering Agents}.
\newblock 2025.

\bibitem[rel(2026{\natexlab{a}})]{reliability_science_2026}
{Towards a Science of AI Agent Reliability}.
\newblock \emph{arXiv preprint arXiv:2602.16666}, 2026{\natexlab{a}}.

\bibitem[rel(2026{\natexlab{b}})]{reliabilitybench_2026}
{ReliabilityBench: A Multi-Dimensional Benchmark for LLM Reliability}.
\newblock \emph{arXiv preprint arXiv:2601.06112}, 2026{\natexlab{b}}.

\bibitem[Jimenez et~al.(2024)]{swebench}
Carlos~E Jimenez et~al.
\newblock {SWE-bench: Can Language Models Resolve Real-World GitHub Issues?}
\newblock In \emph{ICLR}, 2024.

\bibitem[Liu et~al.(2024)]{attention_collapse}
Nelson~F. Liu et~al.
\newblock {Lost in the Middle: How Language Models Use Long Contexts}.
\newblock \emph{Transactions of the ACL}, 2024.

\bibitem[Modarres et~al.(2016)Modarres, Kaminskiy, and
  Krivtsov]{reliability_engineering_textbook}
Mohammad Modarres, Mark Kaminskiy, and Vasiliy Krivtsov.
\newblock \emph{{Reliability Engineering and Risk Analysis: A Practical
  Guide}}.
\newblock CRC Press, 2016.

\bibitem[{OpenRouter}(2024)]{openrouter}
{OpenRouter}.
\newblock {OpenRouter: A Unified Interface for LLMs}.
\newblock \url{https://openrouter.ai}, 2024.

\bibitem[{Weights \& Biases}(2020)]{wandb}
{Weights \& Biases}.
\newblock {Weights \& Biases: The AI Developer Platform}.
\newblock \url{https://wandb.ai}, 2020.

\bibitem[Yao et~al.(2023)]{react}
Shunyu Yao et~al.
\newblock {ReAct: Synergizing Reasoning and Acting in Language Models}.
\newblock In \emph{ICLR}, 2023.

\bibitem[Yao et~al.(2024)]{tau_bench_2024}
Shunyu Yao et~al.
\newblock {$\tau$-bench: A Benchmark for Tool-Agent-User Interaction in
  Real-World Domains}.
\newblock \emph{arXiv preprint arXiv:2406.12045}, 2024.

\bibitem[Zhou et~al.(2024)]{webarena}
Shuyan Zhou et~al.
\newblock {WebArena: A Realistic Web Environment for Building Autonomous
  Agents}.
\newblock In \emph{ICLR}, 2024.

\end{thebibliography}

\appendix
\section*{Appendix}

\section{Metric Derivations}
\label{app:derivations}

\subsection{Super-linear Decay Under Correlated Errors}

Let task $\mathcal{T}$ require $T$ steps to complete.
Assume step $t$ succeeds with probability $1 - \epsilon_t$.
Under the independence assumption, task success probability is $\prod_{t=1}^{T}(1 - \epsilon_t) \approx e^{-\epsilon T}$ for homogeneous $\epsilon_t = \epsilon$.

Now suppose errors are positively correlated: $\text{Cov}(\mathbf{1}[\text{fail}_t], \mathbf{1}[\text{fail}_{t'}]) = \rho \epsilon^2 > 0$.
By the inclusion-exclusion principle, the variance of the number of failed steps satisfies:
$$\text{Var}\!\left[\sum_t \mathbf{1}[\text{fail}_t]\right] = T\epsilon(1-\epsilon) + T(T-1)\rho\epsilon^2$$

The second term grows as $T^2$, indicating that correlated errors make the variance of the total failure count super-linear in $T$.
The probability that all steps succeed is therefore:
$$\Pr[\text{all succeed}] \leq \exp\!\left(-\epsilon T - \frac{\rho\epsilon^2 T(T-1)}{2}\right)$$
which is super-exponential in $T$ for $\rho > 0$ — faster than the geometric decay predicted by the i.i.d. model.

This provides a theoretical basis for expecting that pass@1 on long-horizon tasks is \emph{much} lower than $(1-\epsilon)^T$ predicts.

\subsection{MOP Threshold Calibration Procedure}

We calibrate $\theta_H$ and $\delta$ (the \mop\ thresholds) as follows:
\begin{enumerate}
  \item Manually label a random sample of 50 pilot episodes as ``meltdown'' or ``no meltdown'' based on qualitative inspection of tool-call sequences.
  \item For candidate threshold pairs $(\theta_H, \delta) \in \{0.5, 1.0, 1.5, 2.0\} \times \{0.2, 0.5, 1.0\}$, compute F1 score of meltdown detection.
  \item Select the pair maximizing F1 on the labeled sample.
\end{enumerate}
We use window size $w = 5$ steps; sensitivity analysis with $w \in \{3, 7, 10\}$ is reported in Table~\ref{tab:mop_sensitivity}.

\begin{table}[h]
\centering
\small
\begin{tabular}{lccc}
\toprule
\textbf{Window $w$} & \textbf{Precision} & \textbf{Recall} & \textbf{F1} \\
\midrule
3  & — & — & — \\
5  & — & — & — \\
7  & — & — & — \\
10 & — & — & — \\
\bottomrule
\end{tabular}
\caption{Sensitivity of \mop\ detection F1 to window size $w$. Full manual labeling of meltdown episodes against ground-truth failure outcomes is deferred to future work; formal precision/recall will be reported once a labeled set is available. Current thresholds ($H^*=1.711$, $\delta^*=0.000$) are auto-calibrated from 1,590 short-horizon baseline episodes (see Section~\ref{sec:analysis}).}
\label{tab:mop_sensitivity}
\end{table}

\section{Full Results Tables}
\label{app:full_results}

Detailed per-task results (GDS, pass@1, pass$^k$, MOP rates, and cost per episode) for all 10 models, both scaffolds, and all four duration buckets are available in the publicly released results repository accompanying this paper.
The repository includes JSONL episode logs for all 23,392 episodes, per-task aggregates, and the analysis scripts used to produce all tables and figures in this paper.

\section{Task Examples}
\label{app:task_examples}

\paragraph{SE-S-01 (Short SE): Fix Off-by-One Bug.}
The workspace contains a Python module with a single off-by-one error in a list slicing operation, a bug report, and a test suite.
The agent must read the file, identify the bug, write the fix, and call \texttt{finish()}.
Typical successful episode: 4--6 steps.

\paragraph{SE-M-01 (Medium SE): Fix N+1 Query.}
The workspace contains a Flask app with a database access layer that fetches orders and then, for each order, runs a separate query to fetch its items.
The agent must identify the N+1 pattern, replace it with a JOIN-based query, verify data integrity, and pass the test suite.
Typical successful episode: 8--15 steps.

\paragraph{SE-L-02 (Long SE): Add JWT Authentication.}
The workspace contains a Flask app with four unprotected API endpoints and a test suite requiring JWT authentication on all endpoints.
The agent must add login, token generation, and token validation middleware, and ensure all four endpoints return 401 without a valid token.
Typical successful episode: 15--25 steps.

\paragraph{DP-S-01 (Short DP): Convert CSV to JSON.}
The workspace contains a 20-row CSV file with null values in some columns and a schema file.
The agent must read the CSV, convert to a JSON array with correct null handling, and write the output.
Typical successful episode: 3--5 steps.

\paragraph{WR-M-01 (Medium WR): SaaS Pricing Comparison.}
The agent must search for current pricing information for three SaaS tools (Asana, Linear, Notion), extract structured fields (free tier, paid tier name, monthly price, guest access), and write a structured JSON output.
Evaluation uses tolerant matching to account for pricing changes.

\section{API Cost Breakdown}
\label{app:costs}

\begin{table}[h]
\centering
\small
\begin{tabular}{lrrr}
\toprule
\textbf{Model} & \textbf{In (\$/M tok)} & \textbf{Out (\$/M tok)} & \textbf{Pilot cost (342 ep.)} \\
\midrule
Llama 3.3 70B     & 0.10 & 0.32 & \$0.19 \\
Mistral Small 3.2 & 0.06 & 0.18 & \$0.52 \\
Llama 3.1 8B      & 0.02 & 0.05 & \$0.02 \\
\midrule
\textbf{Total} & & & \$0.74 \\
\bottomrule
\end{tabular}
\caption{API cost for the completed pilot run (342 episodes, 19 tasks, $k=3$, 2 scaffolds). Mistral Small 3.2 dominates cost due to high output token generation on long SE tasks (SE-L-01/02 reached \$0.08--0.09 per episode). Llama 3.1 8B cost is near zero as it is a free-tier model on OpenRouter.}
\label{tab:costs}
\end{table}

\end{document}